\documentclass[runningheads]{llncs}
\usepackage{graphicx}
\usepackage{comment}
\usepackage{amsmath,amssymb} % define this before the line numbering.
\usepackage{color}
\usepackage{epsfig}
\usepackage{xspace}

\usepackage{multirow}
\usepackage{xcolor}
\usepackage{caption} 
\usepackage{pifont}
\def\definedas{\triangleq}
\makeatletter
\DeclareRobustCommand\onedot{\futurelet\@let@token\@onedot}
\def\@onedot{\ifx\@let@token.\else.\null\fi\xspace}
\def\etal{\emph{et al}\onedot}
 
\def\ie{\emph{i.e}\onedot} 

\makeatletter
\def\thickhline{%
  \noalign{\ifnum0=`}\fi\hrule \@height \thickarrayrulewidth \futurelet
   \reserved@a\@xthickhline}
\def\@xthickhline{\ifx\reserved@a\thickhline
               \vskip\doublerulesep
               \vskip-\thickarrayrulewidth
             \fi
      \ifnum0=`{\fi}}
\makeatother

\newlength{\thickarrayrulewidth}
\makeatletter
\def\mediumhline{%
  \noalign{\ifnum0=`}\fi\hrule \@height \mediumarrayrulewidth \futurelet
   \reserved@a\@xthickhline}
\def\@xmediumhline{\ifx\reserved@a\mediumhline
               \vskip\doublerulesep
               \vskip-\mediumarrayrulewidth
             \fi
      \ifnum0=`{\fi}}
\makeatother

\newlength{\mediumarrayrulewidth}
\usepackage{enumitem}
\usepackage[latin1]{inputenc}
\usepackage{subcaption}

\usepackage[breaklinks=true,bookmarks=false]{hyperref}
\usepackage{url}
\hypersetup{
    colorlinks = true,
    urlcolor = blue,
    citecolor=.,
    linkcolor=.,
}

\usepackage{cleveref}

\begin{document}
% \renewcommand\thelinenumber{\color[rgb]{0.2,0.5,0.8}\normalfont\sffamily\scriptsize\arabic{linenumber}\color[rgb]{0,0,0}}
% \renewcommand\makeLineNumber {\hss\thelinenumber\ \hspace{6mm} \rlap{\hskip\textwidth\ \hspace{6.5mm}\thelinenumber}}
% \linenumbers
\pagestyle{headings}
\mainmatter
\def\ECCVSubNumber{2630}  % Insert your submission number here

\title{Probabilistic Future Prediction for\\Video Scene Understanding} % Replace with your title

% INITIAL SUBMISSION 
\begin{comment}
\titlerunning{ECCV-20 submission ID \ECCVSubNumber} 
\authorrunning{ECCV-20 submission ID \ECCVSubNumber} 
\author{Anonymous ECCV submission}
\institute{Paper ID \ECCVSubNumber}
\end{comment}
%******************

% CAMERA READY SUBMISSION
%\begin{comment}
\titlerunning{Probabilistic Future Prediction for Video Scene Understanding}
% If the paper title is too long for the running head, you can set
% an abbreviated paper title here
%
\author{\small{Anthony Hu\inst{1,2} \and
Fergal Cotter\inst{1}\and
Nikhil Mohan\inst{1}\and
Corina Gurau\inst{1}\and
Alex Kendall\inst{1,2}}}
\authorrunning{A. Hu, F. Cotter, N. Mohan, C. Gurau, A. Kendall}
% First names are abbreviated in the running head.
% If there are more than two authors, 'et al.' is used.
%
\institute{Wayve, London, UK. \email{research@wayve.ai}\and
University of Cambridge.}
%\end{comment}
%******************
\maketitle

%%%%%%%%% ABSTRACT
\begin{abstract}
We present a novel deep learning architecture for probabilistic future prediction from video. We predict the future semantics, geometry and motion of complex real-world urban scenes and use this representation to control an autonomous vehicle. This work is the first to jointly predict ego-motion, static scene, and the motion of dynamic agents in a probabilistic manner, which allows sampling consistent, highly probable futures from a compact latent space.
Our model learns a representation from RGB video with a spatio-temporal convolutional module. The learned representation can be explicitly decoded to future semantic segmentation, depth, and optical flow, in addition to being an input to a learnt driving policy.
To model the stochasticity of the future, we introduce a conditional variational approach which minimises the divergence between the present distribution (what could happen given what we have seen) and the future distribution (what we observe actually happens). During inference, diverse futures are generated by sampling from the present distribution.
\end{abstract} 

%%%%%%%%% BODY TEXT
\section{Introduction}

% Building predictive cognitive models of the world is an intrinsically human capability, often regarded as the essence of intelligence.
% It is one of the first skills that we develop as infants, which further enhances our capability to learn more complex tasks, such as navigating or manipulating objects \cite{piaget}.

Building predictive cognitive models of the world is often regarded as the essence of intelligence.
It is one of the first skills that we develop as infants.
We use these models to enhance our capability at learning more complex tasks, such as navigation or manipulating objects \cite{piaget}.

% 1) Problem for AVs
Unlike in humans, developing prediction models for autonomous vehicles to anticipate the future remains hugely challenging.
Road agents have to make reliable decisions based on forward simulation to understand how relevant parts of the scene will evolve.
There are various reasons why modelling the future is incredibly difficult: natural-scene data is rich in details, most of which are irrelevant for the driving task, dynamic agents have complex temporal dynamics, often controlled by unobservable variables, and the future is inherently uncertain, as multiple futures might arise from a unique and deterministic past.

% 2) Current approaches
% we need to draw further comparison to current approaches, which either (1) independently predict each agent on their own, (2) don't consider scene prediction, or (3) have no explicit prediction and rely on representation or q-function.

Current approaches to autonomous driving individually model each dynamic agent by producing hand-crafted behaviours, such as trajectory forecasting, to feed into a decision making module \cite{intentnet18}. 
This largely assumes independence between agents and fails to model multi-agent interaction.  % actually most works assume independence between agents or model "robot-only" influence
Most works that holistically reason about the temporal scene are limited to simple, often simulated environments or use low dimensional input images that do not have the visual complexity of real world driving scenes \cite{video_pred_atari15}.
Some approaches tackle this problem by making simplifying assumptions to the motion model or the stochasticity of the world \cite{NIPS2014_5444,intentnet18}.
Others avoid explicitly predicting the future scene but rather rely on an implicit representation or Q-function (in the case of model-free reinforcement learning) in order to choose an action \cite{model_free_rl_ex1,model_free_rl_ex2,kendall2019learning}.
% "the robot freezing problem" - future becomes too uncertain to operate (Andres Krause)
%(3) Motion, why modelling motion/video representations is hard and what we propose to do this
% In this work we describe a deep learning architecture for representing the dynamics from video and jointly predicting ego-motion, static scene representation and the motion of dynamic agents, in a probabilistic manner.

% (5) Multimodality, why probabilistic modelling is so important and how we do it
Real world future scenarios are difficult to model because of the stochasticity and the partial observability of the world.
Our work addresses this by encoding the future state into a low-dimensional \textit{future distribution}. We then allow the model to have a privileged view of the future through the future distribution at training time. As we cannot use the future at test time, we train a \emph{present distribution} (using only the current state) to match the future distribution through a Kullback-Leibler (KL) divergence loss. We can then sample from the present distribution during inference, when we do not have access to the future.
%Our work addresses this by using an approximate \textit{future distribution} model  which receives as input the present state as well as the future scene understanding representation (in the form of semantic segmentation, depth and flow).
%At training time a latent value is sampled from this distribution and passed as input to the dynamics model.
%At inference time we sample from another distribution (\textit{the present distribution}) which has been trained to match the future distribution through a Kullback-Leibler (KL) divergence loss.
We observe that this paradigm allows the model to learn accurate and diverse probabilistic future prediction outputs.
% Through data balancing over the action space, we show the model more interesting driving sequences from which it learns to produce a multimodal output.
%  Or say this somehow: As the model will see different outcomes from the same sequence of inputs, it will learn to model all the modes in its prior distribution.

In order to predict the future we need to first encode video into a motion representation.
Unlike advances in 2D convolutional architectures \cite{inception14,he16}, learning spatio-temporal features is more challenging due to the higher dimensionality of video data and the complexity of modelling dynamics.
State-of-the-art architectures \cite{chen18,tran18} decompose 3D filters into spatial and temporal convolutions in order to learn more efficiently.
The model we propose further breaks down convolutions into many space-time combinations and context aggregation modules, stacking them together in a more complex hierarchical representation.
We show that the learnt representation is able to jointly predict ego-motion and motion of other dynamic agents.
By explicitly modelling these dynamics we can capture the essential features for representing causal effects for driving. Ultimately we use this motion-aware and future-aware representation to improve an autonomous vehicle control policy.

% We show that by having a stochastic model we further improve both perception and control.

%  (6) Brief summary of results and contributions
Our main contributions are threefold.
Firstly, we present a novel deep learning framework for future video prediction.
Secondly, we demonstrate that our probabilistic model is able to generate visually diverse and plausible futures.
Thirdly, we show our future prediction representation substantially improves a learned autonomous driving policy.

\section{Related Work}

%%%%%%%%%%%%%
This work falls in the intersection of learning scene representation from video, probabilistic modelling of the ambiguity inherent in real-world driving data, and using the learnt representation for control.

\paragraph{Temporal representations.} 
Current state-of-the-art temporal representations from video use recurrent neural networks \cite{shi15,siam2017convolutional}, separable 3D convolutions \cite{Ioannou2016,SunJYS15,XieGDTH16,hara17,tran18}, or 3D Inception modules \cite{carreira17,chen18}. 
In particular, the separable 3D Inception (S3D) architecture \cite{chen18}, which improves on the Inception 3D module (I3D) introduced by Carreira \etal \cite{carreira17}, shows the best trade-off between model complexity and speed, both at training and inference time.
Adding optical flow as a complementary input modality has been consistently shown to improve performance \cite{Feichtenhofer16,simonyan14,simonyan15,Bilen2016DynamicIN}, in particular using flow for representation warping to align features over time \cite{gadde2017semantic,zhu2017deep}.
We propose a new spatio-temporal architecture that can learn hierarchically more complex features with a novel 3D convolutional structure incorporating both local and global space and time context. 

\paragraph{Visual prediction.}
Most works for learning dynamics from video fall under the framework of model-based reinforcement learning \cite{finn16a,finn17,Ebert18,kaizer19} or unsupervised feature learning \cite{srivastava15,denton17}, both regressing directly in pixel space \cite{mathieu16,ranzato14,Jayaraman19} or in a learned feature space \cite{Jaderberg16,finn16}.
For the purpose of creating good representations for driving scenes, directly predicting in the high-dimensional space of image pixels is unnecessary, as some details about the appearance of the world are irrelevant for planning and control.
Our approach is similar to that of Luc \etal~\cite{luc17} which trains a model to predict future semantic segmentation using pseudo-ground truth labels generated from a teacher model. However, our model predicts a more complete scene representation with segmentation, depth, and flow and is probabilistic in order to model the uncertainty of the future.

\paragraph{Multi-modality of future prediction.} 
Modelling uncertainty is important given the stochastic nature of real-world data \cite{kendall2017uncertainties}.
Lee \etal \cite{desire17}, Bhattacharyya \etal \cite{Bhattacharyya18} and Rhinehart \etal \cite{precog19} forecast the behaviour of other dynamic agents in the scene in a probabilistic multi-modal way. 
We distinguish ourselves from this line of work as their approach does not consider the task of video forecasting, but rather trajectory forecasting, and they do not study how useful the representations learnt are for robot control.
Kurutach \etal~\cite{Kurutach2018LearningPR} propose generating multi-modal futures with adversarial training, however spatio-temporal discriminator networks are known to suffer from mode collapse \cite{goodfellow2016tutorial}.

Our variational approach is similar to Kohl \etal \cite{kohl18}, although their application domain does not involve modelling dynamics.
Furthermore, while Kohl \etal \cite{kohl18} use multi-modal training data, i.e. multiple output labels are provided for a given input, we learn directly from real-world driving data, where we can only observe one future reality, and show that we generate diverse and plausible futures. Most importantly, previous variational video generation methods \cite{lee18,denton18} were restricted to single-frame image generation, low resolution ($64\times64$) datasets that are either simulated (Moving MNIST \cite{srivastava15}) or with static scenes and limited dynamics (KTH actions \cite{schuldt04}, Robot Pushing dataset \cite{ebert17}). Our new framework for future prediction generates entire video sequences on complex real-world urban driving data with ego-motion and complex interactions.

\paragraph{Learning a control policy.}
The representation learned from dynamics models could be used to generate imagined experience to train a policy in a model-based reinforcement learning setting \cite{ha18,hafner2019planet} or to run shooting methods for planning \cite{Chua18}.
Instead we follow the approaches of Bojarski \etal~\cite{bojarski_end_2016}, Codevilla \etal~\cite{codevilla2018end} and Amini \etal~\cite{amini2018variational} and learn a policy which predicts longitudinal and lateral control of an autonomous vehicle using Conditional Imitation Learning, as this approach has been shown to be immediately transferable to the real world. 

%%%%%%%%%%%%%%%%%%%%%%%%%%%%%%%%%%%%%%%%%%%%%%%%%%%%%%%%%%%%%%%%%%%%%%%%%%%%%%%%%%%%%%%%%%%%%%%%%%%%%%%%%%%%%%%%%
% ---------------------------------------------------------------------------------------------------------------
%%%%%%%%%%%%%%%%%%%%%%%%%%%%%%%%%%%%%%%%%%%%%%%%%%%%%%%%%%%%%%%%%%%%%%%%%%%%%%%%%%%%%%%%%%%%%%%%%%%%%%%%%%%%%%%%%

\begin{figure*}[t!]
  \centering
  \includegraphics[width=\linewidth]{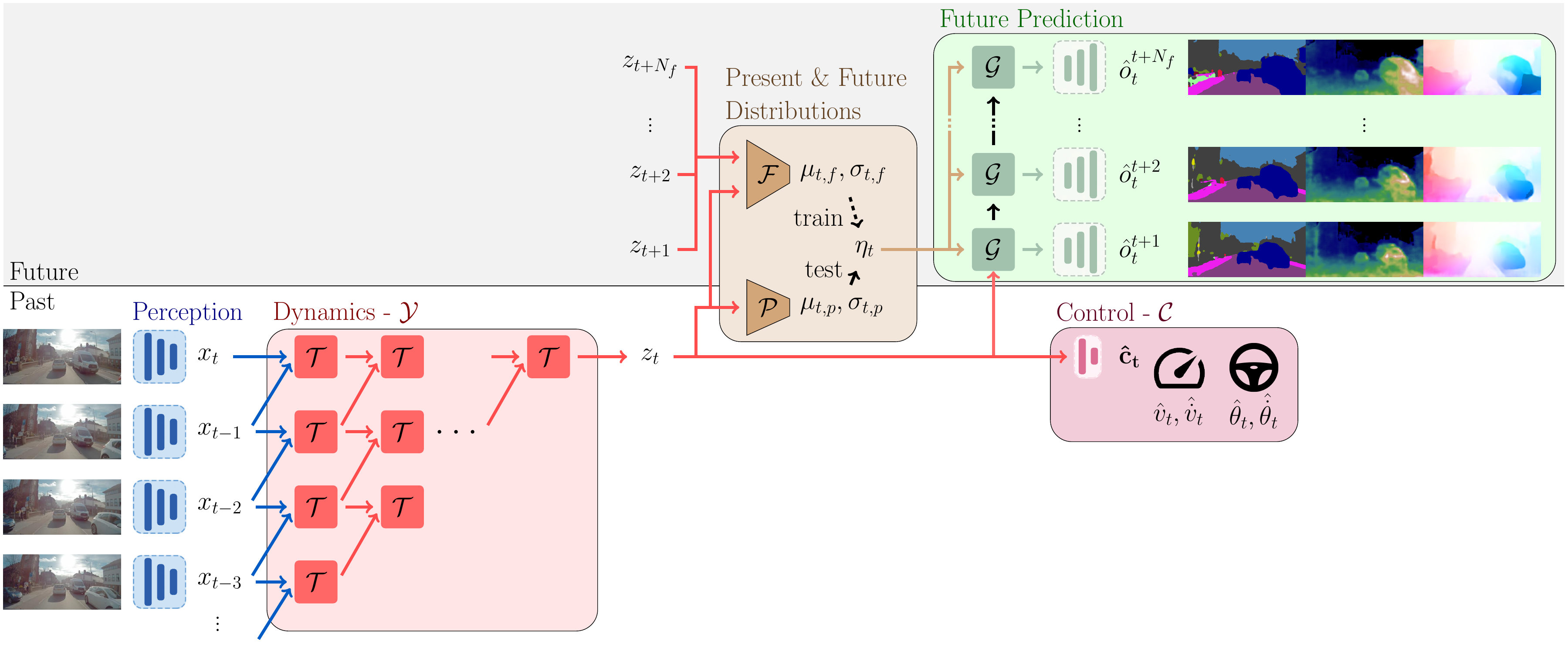}
  \caption{Our architecture has five modules: \textit{Perception}, \textit{Dynamics}, \textit{Present/Future Distributions}, \textit{Future Prediction} and \textit{Control}. The Perception module learns scene representation features, $x_t$, from input images. The Dynamics model builds on these scene features to produce a spatio-temporal representation, $z_t$, with our proposed \textit{Temporal Block} module, $\mathcal{T}$. Together with a noise vector, $\eta_t$, sampled from a future distribution, $\mathcal{F}$, at training time, or the present distribution, $\mathcal{P}$, at inference time, this representation predicts future video scene representation (segmentation, depth and optical flow) with a convolutional recurrent model, $\mathcal{G}$, and decoders, $\mathcal{D}$. Lastly, we learn a Control policy, $\mathcal{C}$, from the spatio-temporal representation, $z_t$.}
  \label{fig:architecture}
\end{figure*}

\section{Model Architecture}
Our model learns a spatio-temporal feature to jointly predict future scene representation (semantic segmentation, depth, optical flow) and train a driving policy. The architecture contains five components: \textit{Perception}, an image scene understanding model, \textit{Dynamics}, which learns a spatio-temporal representation, \textit{Present/Future Distributions}, our probabilistic framework, \textit{Future Prediction}, which predicts future video scene representation, and \textit{Control}, which trains a driving policy using expert driving demonstrations.
\autoref{fig:architecture} gives an overview of the model and further details are described in this section and \Cref{appendix:arch}.

\subsection{Perception}
The perception component of our system contains two modules: the encoder of a scene understanding model that was trained on single image frames to reconstruct semantic segmentation and depth \cite{kendall2017multi}, and the encoder of a flow network \cite{sun_pwc-net_2018}, trained to predict optical flow. The combined perception features $x_t \in \mathbb{R}^{C\times H \times W}$ form the input to the dynamics model. These models can also be used as a teacher to distill the information from the future, giving pseudo-ground truth labels for segmentation, depth and flow $\{s_t, d_t, f_t\}$. See \autoref{sec:training_data} for more details on the teacher model.

\subsection{Dynamics}
Learning a temporal representation from video is extremely challenging because of the high dimensionality of the data, the stochasticity and complexity of natural scenes, and the partial observability of the environment. To train 3D convolutional filters from a sequence of raw RGB images, a large amount of data, memory and compute is required. We instead learn spatio-temporal features with a temporal model that operates on perception encodings, which constitute a more powerful and compact representation compared to RGB images.

The dynamics model $\mathcal{Y}$ takes a history of perception features $(x_{t-T+1}:x_t)$ with temporal context $T$ and encodes it into a dynamics feature $z_t$:
\begin{equation}\label{eq:dynamics}
    z_t = \mathcal{Y}(x_{t-T+1}:x_t) 
\end{equation}

\paragraph{Temporal Block}
We propose a spatio-temporal module, named {\emph{Temporal Block}}, to learn hierarchically more complex temporal features as follows:

\begin{itemize}[noitemsep,topsep=0pt,parsep=2pt,partopsep=2pt]
    \item \textbf{Decomposing the filters}: instead of systematically using full 3D filters $(k_t, k_s, k_s)$, with $k_t$ the time kernel dimension and $k_s$ the spatial kernel dimension, we apply four parallel 3D convolutions with kernel sizes: $(1, k_s, k_s)$ (spatial features), $(k_t, 1, k_s)$ (horizontal motion), $(k_t, k_s, 1)$ (vertical motion), and $(k_t, k_s, k_s)$ (complete motion). All convolutions are preceded by a $(1, 1, 1)$ convolution to compress the channel dimension.
    \item \textbf{Global spatio-temporal context}: in order to learn contextual features, we additionally use three spatio-temporal average pooling layers at: full spatial size $(k_t, H, W)$ ($H$ and $W$ are respectively the height and width of the perception features $x_t$), half size $(k_t, \frac{H}{2}, \frac{W}{2})$ and quarter size  $(k_t, \frac{H}{4}, \frac{W}{4})$, followed by bilinear upsampling to the original spatial dimension $(H, W)$ and a $(1, 1, 1)$ convolution. 
\end{itemize}

\begin{figure}[t]
    \centering
    \includegraphics[width=.6\textwidth]{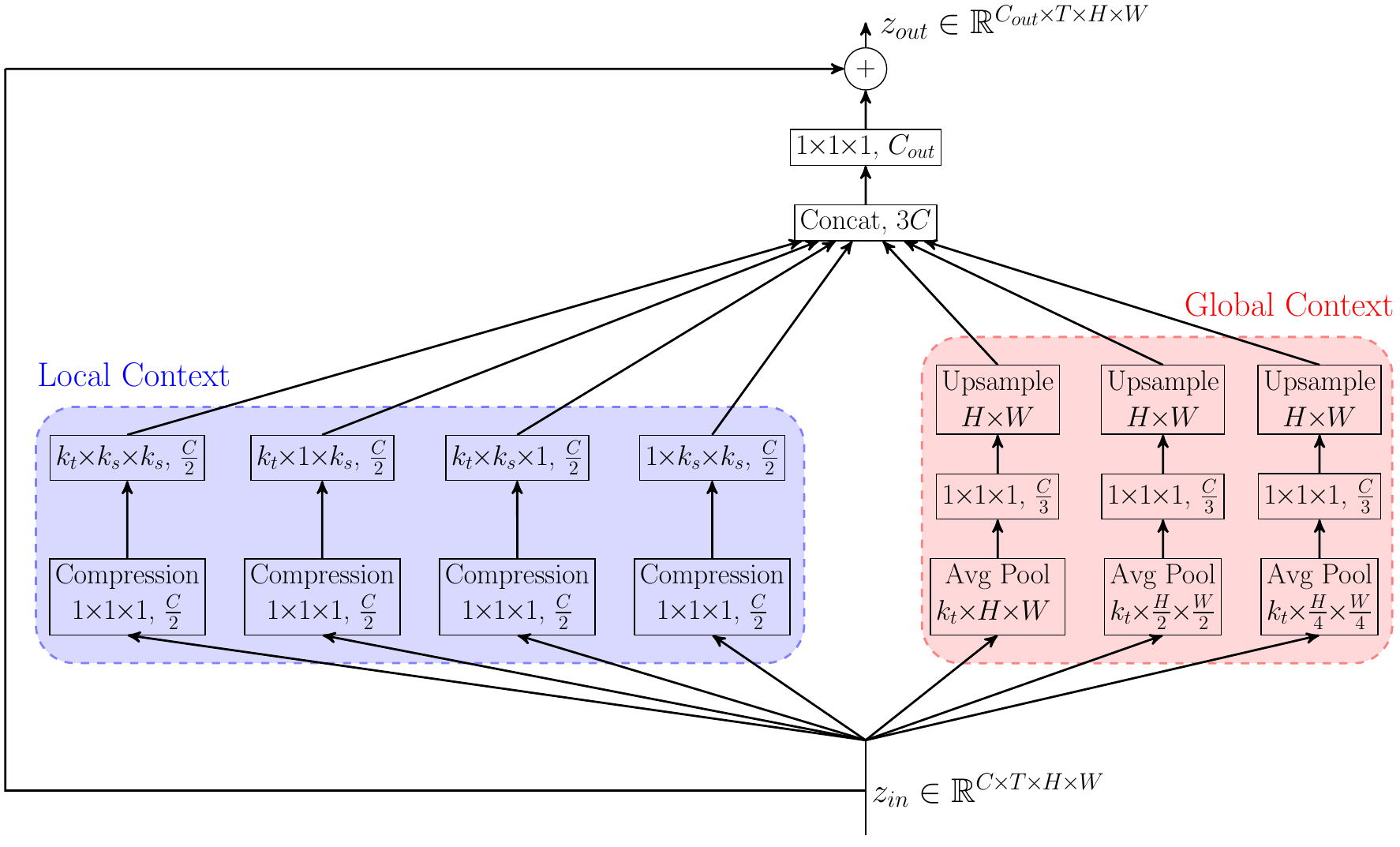}
    \caption{A Temporal Block, our proposed spatio-temporal module. From a four-dimensional input $z_{in} \in \mathbb{R}^{C\times T \times H \times W}$, our module learns both local and global spatio-temporal features. The local head learns all possible configurations of 3D convolutions with filters: $(1, k_s, k_s)$ (spatial features), $(k_t, 1, k_s)$ (horizontal motion), $(k_t, k_s, 1)$ (vertical motion), and $(k_t, k_s, k_s)$ (complete motion). The global head learns global spatio-temporal features with a 3D average pooling at full, half and quarter size, followed by a $(1, 1, 1)$ convolution and upsampling to the original spatial dimension $H\times W$. The local and global features are then concatenated and combined in a final $(1, 1, 1)$ 3D convolution.}
    \label{fig:temporal_block}
\end{figure}

\autoref{fig:temporal_block} illustrates the architecture of the Temporal Block. By stacking multiple temporal blocks, the network learns a representation that incorporates increasingly more temporal, spatial and global context. We also increase the number of channels by a constant $\alpha$ after each temporal block, as after each block, the network has to represent the content of the $k_t$ previous features.

%As time is a special dimension \cite{carreira17}, stacking temporal blocks one after the other would be sub-efficient since the network has to learn a joint representation of the $k_t$ previous features. To give the network more representational capacity, we interleave the temporal blocks with spatial residual convolutions \cite{he16}. 

\subsection{Future Prediction}
%The dynamics feature $z_t$ is a compact scene representation of the past context, and as such, it contains enough information to predict the future.
We train a future prediction model that unrolls the dynamics feature, which is a compact scene representation of the past context, into predictions about the state of the world in the future. The future prediction model is a convolutional recurrent network $\mathcal{G}$ which creates future features $g_t^{t+i}$ that become the inputs of individual decoders $\mathcal{D}_s, \mathcal{D}_d, \mathcal{D}_f$ to decode these features to predicted segmentation $\hat{s}_t^{t+i}$, depth $\hat{d}_t^{t+i}$, and flow $\hat{f}_t^{t+i}$ values in the pixel space. We have introduced a second time superscript notation, \ie $g_t^{t+i}$, represents the prediction about the world at time $t+i$ given the dynamics features at time $t$. Also note that $g_t^{t} \definedas z_t$.

The structure of the convolutional recurrent network $\mathcal{G}$ is the following: a convolutional GRU \cite{ballas16} followed by three spatial residual layers, repeated $D$ times, similarly to Clark \etal \cite{clark19}.  For deterministic inference, its input is $u_t^{t+i} = \mathbf{0}$, and its initial hidden state is $z_t$, the dynamics feature. The future prediction component of our network computes the following, for $i \in \{1,.., N_f\}$, with $N_f$ the number of predicted future frames:
\begin{align}
    g_t^{t+i} &= \mathcal{G}(u_t^{t+i}, g_t^{t+i-1}) \\
    \hat{s}_t^{t+i} &= \mathcal{D}_s(g_t^{t+i}) \\
    \hat{d}_t^{t+i} &= \mathcal{D}_d(g_t^{t+i}) \\
    \hat{f}_t^{t+i} &= \mathcal{D}_f(g_t^{t+i}) 
\end{align}

%----------------------------------------------------
\subsection{Present \& Future Distributions}
From a unique past in the real-world, many futures are possible, but in reality we only observe one future. Consequently, modelling multi-modal futures from deterministic video training data is extremely challenging. We adopt a conditional variational approach and model two probability distributions: a {\emph{present distribution}} $P$, that represents what could happen given the past context, and a {\emph{future distribution}} $F$, that represents what actually happened in that particular observation. This allows us to learn a multi-modal distribution from the input data while conditioning the model to learn from the specific observed future from within this distribution.

The present and the future distributions are diagonal Gaussian, and can therefore be fully characterised by their mean and standard deviation. We parameterise both distributions with a neural network, respectively $\mathcal{P}$ and $\mathcal{F}$.

\paragraph{Present distribution} The input of the network $\mathcal{P}$ is $z_t \in \mathbb{R}^{C_d \times H \times W}$, which represents the past context of the last $T$ frames ($T$ is the time receptive field of our dynamics module). The present network contains two downsampling convolutional layers, an average pooling layer and a fully connected layer to map the features to the desired latent dimension $L$. The output of the network is the parametrisation of the present distribution: $(\mu_{t,\text{present}}, \sigma_{t,\text{present}}) \in \mathbb{R}^L \times \mathbb{R}^L$.

\paragraph{Future distribution} $\mathcal{F}$ is not only conditioned by the past $z_t$, but also by the future corresponding to the training sequence. Since we are predicting $N_f$ steps in the future, the input of $\mathcal{F}$ has to contain information about future frames $(t+1, ..., t+N_f)$. This is achieved using the learned dynamics features $\{z_{t + j}\}_{j \in J}$, with $J$ the set of indices such that $\{z_{t + j}\}_{j \in J}$ covers all future frames $(t+1, ..., t+N_f)$, as well as $z_t$. Formally, if we want to cover $N_f$ frames with features that have a receptive field of $T$, then: \\ $J = \{nT ~ |~ 0 \le n \le \lfloor N_f / T \rfloor \} \cup \{N_f\}$.
The architecture of the future network is similar to the present network: for each input dynamics feature $z_{t+j} \in \mathbb{R}^{C_d \times H \times W}$, with $j \in F$, we apply two downsampling convolutional layers and an average pooling layer. The resulting features are concatenated, and a fully-connected layer outputs the parametrisation of the future distribution: 
\\$(\mu_{t,\text{future}}, \sigma_{t,\text{future}}) \in \mathbb{R}^L \times \mathbb{R}^L$.

\paragraph{Probabilistic Future Prediction}
During training, we sample from the future distribution a vector $\eta_t \sim \mathcal{N}(\mu_{t,\text{future}}, \sigma_{t,\text{future}}^2)$ that conditions the predicted future perception outputs (semantic segmentation, depth, optical flow) on the observed future. As we want our prediction to be consistent in both space and time, we broadcast spatially $\eta_t \in \mathbb{R}^L$ to $\mathbb{R}^{L\times H \times W}$, and use the same sample throughout the future generation as an input to the GRU to condition the future: for $i \in \{1,..,N_f\}$, input $u_t^{t+i} = \eta_t$.

We encourage the present distribution $P$ to match the future distribution $F$ with a mode-covering KL loss:
\begin{equation}
    L_{\text{probabilistic}} = D_\text{KL}(F(\cdot| Z_t,..., Z_{t+N_f}) ~ || ~P(\cdot | Z_t))
\end{equation}
As the future is multimodal, different futures might arise from a unique past context $z_t$. Each of these futures will be captured by the future distribution $F$ that will pull the present distribution $P$ towards it. Since our training data is extremely diverse, it naturally contains multimodalities. Even if the past context (sequence of images $(i_1, ..., i_t)$) from two different training sequences will never be the same, the dynamics network will have learned a more abstract spatio-temporal representation that ignores irrelevant details of the scene (such as vehicle colour, weather, road material etc.) to match similar past context to a similar $z_t$. In this process, the present distribution will learn to cover all the possible modes contained in the future.

During inference, we sample a vector $\eta_t$ from the present distribution $\eta_t \sim \mathcal{N}(\mu_{t,\text{present}}, \sigma_{t,\text{present}}^2)$, where each sample corresponds to a different future. 

\subsection{Control}
From this rich spatio-temporal representation $z_t$ explicitly trained to predict the future, we train a control model $\mathcal{C}$ to output a four dimensional vector consisting of estimated speed $\hat{v}$, acceleration $\hat{\dot{v}}$, steering angle $\hat{\theta}$ and angular velocity $\hat{\dot{\theta}}$:
\begin{equation}
    \hat{c}_t = \{\hat{v}_t, \hat{\dot{v}}_t, \hat{\theta}_t, \hat{\dot{\theta}}_t\} = \mathcal{C}(z_t)
\end{equation}
$\mathcal{C}$ compresses $z_t \in \mathbb{R}^{C_d \times H \times W}$ with strided convolutional layers, then stacks several fully connected layers, compressing at each stage, to regress the four dimensional output.

\subsection{Losses}

\paragraph{Future Prediction}
The future prediction loss at timestep $t$ is the weighted sum of future segmentation, depth and optical flow losses. Let the segmentation loss at the future timestep $t+i$ be $L_s^{t+i}$. We use a top-k cross-entropy loss \cite{Wu2016BridgingCA} between the network output $\hat{s}_t^{t+i}$ and the pseudo-ground truth label $s_{t+i}$. $L_s$ is computed by summing these individual terms over the future horizon $N_f$ with a weighted discount term $0 < \gamma_f< 1$:
\begin{equation}
    L_s = \sum_{i=0}^{N_f-1} \gamma_f^i L_s^{t+i}
\end{equation}
For depth, $L_d^{t+i}$ is the scale-invariant depth loss \cite{li_megadepth_2018} between $\hat{d}_t^{t+i}$ and $d_{t+i}$, and similarly $L_d$ is the discounted sum. For flow, we use a Huber loss betwen $\hat{f}_t^{t+i}$ and $f_{t+i}$. We weight the summed losses by factors $\lambda_s, \lambda_d, \lambda_f$ to get the future prediction loss $L_{\text{future-pred}}$.
\begin{equation}
    L_{\text{future-pred}} =  \lambda_s L_{s} + \lambda_d L_{d} + \lambda_f L_{f}
\end{equation}

\paragraph{Control}
We use imitation learning, regressing to the expert's true control actions $\{v, \theta \}$ to generate a \emph{control loss} $L_c$. 
For both speed and steering, we have access to the expert actions.

We compare to the linear extrapolation of the generated policy's speed/steering for future time-steps up to $N_c$ frames in the future:
\begin{align}
    L_c = \sum_{i=0}^{N_c-1} \gamma_c^i  & \left(  \left(  v_{t+i} - \left(\hat{v}_t + i \hat{\dot{v}}_t\right) \right)^2 + \right. \nonumber \\ 
    & \hphantom{a.} \left. \left(  \theta_{t+i} - \left(\hat{\theta}_t + i\hat{\dot{\theta}}_t\right) \right)^2 \right)
\end{align}
where $0 < \gamma_c < 1$ is the control discount factor penalizing less speed and steering errors further into the future.
\paragraph{Total Loss}
The final loss $L$ can be decomposed into the future prediction loss ($L_{\text{future-pred}}$), the probabilistic loss ($L_{\text{probabilistic}}$), and the control loss ($L_c$) .
\begin{equation}
    L = \lambda_{fp}{L_\text{future-pred}} + \lambda_c L_c + \lambda_p L_{\text{probabilistic}}
\end{equation}
In all experiments we use $\gamma_f=0.6$, $\lambda_s = 1.0$, $\lambda_d =1.0$, $\lambda_f = 0.5$, $\lambda_{fp}=1$, $\lambda_p=0.005$, $\gamma_c=0.7$, $\lambda_c=1.0$.

\section{Experiments}

We have collected driving data in a densely populated, urban environment, representative of most European cities using multiple drivers over the span of six months.
For the purpose of this work, only the front-facing camera images $i_t$ and the measurements of the speed and steering $c_t$ have been used to train our model, all sampled at $5$Hz.

\subsection{Training Data}
\label{sec:training_data}
\paragraph{Perception} We first pretrain the scene understanding encoder on a number of heterogeneous datasets to predict semantic segmentation and depth: CityScapes \cite{cityscapes16}, Mapillary Vistas \cite{neuhold_mapillary_2017}, ApolloScape \cite{huang2018apolloscape} and Berkeley Deep Drive \cite{yu2018bdd100k}. The optical flow network is a pretrained PWC-Net from \cite{sun_pwc-net_2018}.
The decoders of these networks are used for generating pseudo-ground truth segmentation and depth labels to train our dynamics and future prediction modules.
\paragraph{Dynamics and Control}
The dynamics and control modules are trained using 30 hours of driving data from the urban driving dataset we collected and described above.
We address the inherent dataset bias by sampling data uniformly across lateral and longitudinal dimensions.
First, the data is split into a histogram of bins by steering, and subsequently by speed.
We found that weighting each data point proportionally to the width of the bin it belongs to avoids the need for alternative approaches such as data augmentation.

\subsection{Metrics}
\label{sec:metrics}

We report standard metrics for measuring the quality of segmentation, depth and flow: respectively intersection-over-union, scale-invariant logarithmic error, and average end-point error.
For ease of comparison, additionally to individual metrics, we report a unified perception metric $\mathcal{M}_{\text{perception}}$ defined as improvement of segmentation, depth and flow metrics with respect to the \textit{Repeat Frame} baseline (repeats the perception outputs of the current frame):
\begin{equation}
    \mathcal{M}_{\text{perception}} = \frac{1}{3} (\text{seg}_{\text{\% increase}} + \text{depth}_{\text{\% decrease}} + \text{flow}_{\text{\% decrease}})
\end{equation}
Inspired by the energy functions used in \cite{BellemareDDMLHM17,Salimans18}, we additionally report a \textit{diversity distance metric} $\text{(DDM)}$ between the ground truth future $Y$ and samples from the predicted present distribution $P$: 
\begin{equation}
    \text{DDM}(Y, P)=\min_{S} \big[ d(Y,S) \big] - \mathbb{E} \big[ d(S,S') \big]
\end{equation}
where $d$ is an error metric and $S$, $S'$, are independent samples from the present distribution $P$.
This metric measures performance both in terms of accuracy, by looking at the minimum error of the samples, as well as the diversity of the predictions by taking the expectation of the distance between $N$ samples.
The distance $d$ is the scale-invariant logarithmic error for depth, the average end-point error for flow, and for segmentation $ d(x, y) = 1 - \text{IoU}(x, y)$.

To measure control performance, we report mean absolute error of speed and steering outputs, balanced by steering histogram bins.

\section{Results}
We first compare our proposed spatio-temporal module to previous state-of-the-art architectures and show that our module achieves the best performance on future prediction metrics. Then we demonstrate that modelling the future in a probabilistic manner further improves performance. And finally, we show that our probabilistic future prediction representation substantially improves a learned driving policy. All the reported results are evaluated on test routes with no overlap with the training data.

\subsection{Spatio-Temporal Representation}
We analyse the quality of the spatio-temporal representation our temporal model learns by evaluating future prediction of semantic segmentation, depth, and optical flow, two seconds in the future. Several architectures have been created to learn features from video, with the most successful modules being: the Convolutional GRU \cite{ballas16}, the 3D Residual Convolution \cite{hara17} and the Separable 3D Inception block \cite{chen18}.

We also compare our model to two baselines: \textit{Repeat frame} (repeating the perception outputs of the current frame at time $t$ for each future frame $t+i$ with $i=1,...,N_f$), and \textit{Static} (without a temporal model). As shown in \autoref{table:future-prediction}, deterministic section, every temporal model architecture improves over the \textit{Repeat frame} baseline, as opposed to the model without any temporal context (\textit{Static}), that performs notably worse. This is because it is too difficult to forecast how the future is going to evolve with a single image. 

\begin{table*}[thb!]
\centering
\begin{tabular}{ll|cccc}
\thickhline
& {\textbf{\color{darkgray}{\shortstack{Temporal Model}}}} & $\mathcal{M}_{\text{perception}} (\uparrow$) & Depth ($\downarrow$) & Flow ($\downarrow$) & Seg. ($\uparrow$)\\
\mediumhline
& Repeat frame & 0.0\%  & 1.467 & 5.707& 0.356\\
& Static & -40.3\% & 1.980 & 8.573 & 0.229\\
\hline
\multirow{4}{*}{\textbf{\color{darkgray}{\shortstack{Deterministic \quad}}}}&  Res. 3D Conv. \cite{hara17} & 6.9\% & 1.162& 5.437& 0.339\\
& Conv. GRU \cite{ballas16} & 7.4\%  & 1.097& 5.714& 0.346\\
& Sep. Inception \cite{chen18}& 9.6\% & 1.101 & 5.300& 0.344\\
& \textbf{Ours} & \textbf{13.6\%} & \textbf{1.090} & \textbf{5.029} & \textbf{0.367}\\
\hline
\multirow{4}{*}{\textbf{\color{darkgray}{\shortstack{Probabilistic \quad}}}} & Res. 3D Conv. \cite{hara17} & 8.1\%  & 1.107 & 5.720 & 0.356\\
& Conv. GRU \cite{ballas16} & 9.0\% & 1.101 & 5.645 & 0.359\\
& Sep. Inception \cite{chen18}& 13.8\% & 1.040 & 5.242 & 0.371\\
& \textbf{Ours} & \textbf{20.0\%} & \textbf{0.970} & \textbf{4.857} & \textbf{0.396}\\
\thickhline
\end{tabular}
\caption{Perception performance metrics for two seconds future prediction on the collected urban driving data. We measure semantic segmentation with mean IoU, depth with scale-invariant logarithmic error, and depth with average end-point error. $\mathcal{M}_{\text{perception}}$ shows overall performance --- we observe our model outperforms all baselines.}
\label{table:future-prediction}
\vspace{-20pt}
\end{table*}

Further, we observe that our proposed temporal block module outperforms all preexisting spatio-temporal architectures, on all three future perception metrics: semantic segmentation, depth and flow. There are two reasons for this: the first one is that learning 3D filters is hard, and as demonstrated by the Separable 3D convolution \cite{chen18} (i.e. the succession of a $(1, k_s, k_s)$ spatial filter and a $(k_t, 1, 1)$ time filter), decomposing into two subtasks helps the network learn more efficiently. In the same spirit, we decompose the spatio-temporal convolutions into all combinations of space-time convolutions: $(1, k_s, k_s)$, $(k_t, 1, k_s)$, $(k_t, k_s, 1)$, $(k_t, k_s, k_s)$, and by stacking these temporal blocks together, the network can learn a hierarchically more complex representation of the scene.
The second reason is that we incorporate global context in our features. By pooling the features spatially and temporally at different scales, each individual feature map also has information about the global scene context, which helps in ambiguous situations. \Cref{appendix:temporal-block} contains an ablation study of the different component of the Temporal Block.

\subsection{Probabilistic Future}
Since the future is inherently uncertain, the deterministic model is training in a chaotic learning space because the predictions of the model are penalised with the ground truth future, which only represents a subset of all the possible outcomes. Therefore, if the network predicts a plausible future, but one that did not match the given training sequence, it will be heavily penalised. On the other hand, the probabilistic model has a very clean learning signal as the future distribution conditions the network to generate the correct future. The present distribution is encouraged to match the distribution of the future distribution during training, and therefore has to capture all the modes of the future.

During inference, samples $\eta_t \sim \mathcal{N}(\mu_{t,\text{present}}, \sigma_{t,\text{present}}^2)$ from the present distribution should give a different outcome, with $p(\eta_t | \mu_{t,\text{present}}, \sigma_{t,\text{present}}^2)$ indicating the relative likelihood of a given scenario. Our probabilistic model should be accurate, that is to say at least one of the generated future should match the ground truth future. It should also be diverse: the generated samples should capture the diversity of the possible futures with the correct probability. Next, we analyse quantitatively and qualitatively that our model generates diverse and accurate futures.

\begin{table}[thb!]
\centering
\begin{tabular}{l|ccc}
\thickhline
{\textbf{\color{darkgray}{\shortstack{Temporal Model}}}} & Depth ($\downarrow$) & Flow ($\downarrow$) & Seg. ($\downarrow$)\\ 
\mediumhline
Res. 3D Conv. \cite{hara17} &0.823 & 2.695 &0.474 \\
Conv. GRU \cite{ballas16} & 0.841& 2.683& 0.493\\
Sep. Inception \cite{chen18}& 0.799 & 2.914&0.469\\
\textbf{Ours} & \textbf{0.724} & \textbf{2.676} & \textbf{0.424}\\
\thickhline
\end{tabular}
\caption{Diversity Distance Metric for various temporal models evaluated on the urban driving data, demonstrating that our model produces the most accurate and diverse distribution.}
\label{table:diversity}
\vspace{-16pt}
\end{table}

\autoref{table:future-prediction} shows that every temporal architecture have superior performance when trained in a probabilistic way, with our model benefiting the most (from 13.6\% to 20.0\%) in future prediction metrics.
\autoref{table:diversity} shows that our model outperforms other temporal representations also using the diversity distance metric \text{(DDM)} described in \autoref{sec:metrics}. The \text{DDM} measures both accuracy and diversity of the distribution.

\begin{figure*}[h]
    \centering
    \includegraphics[width=\textwidth]{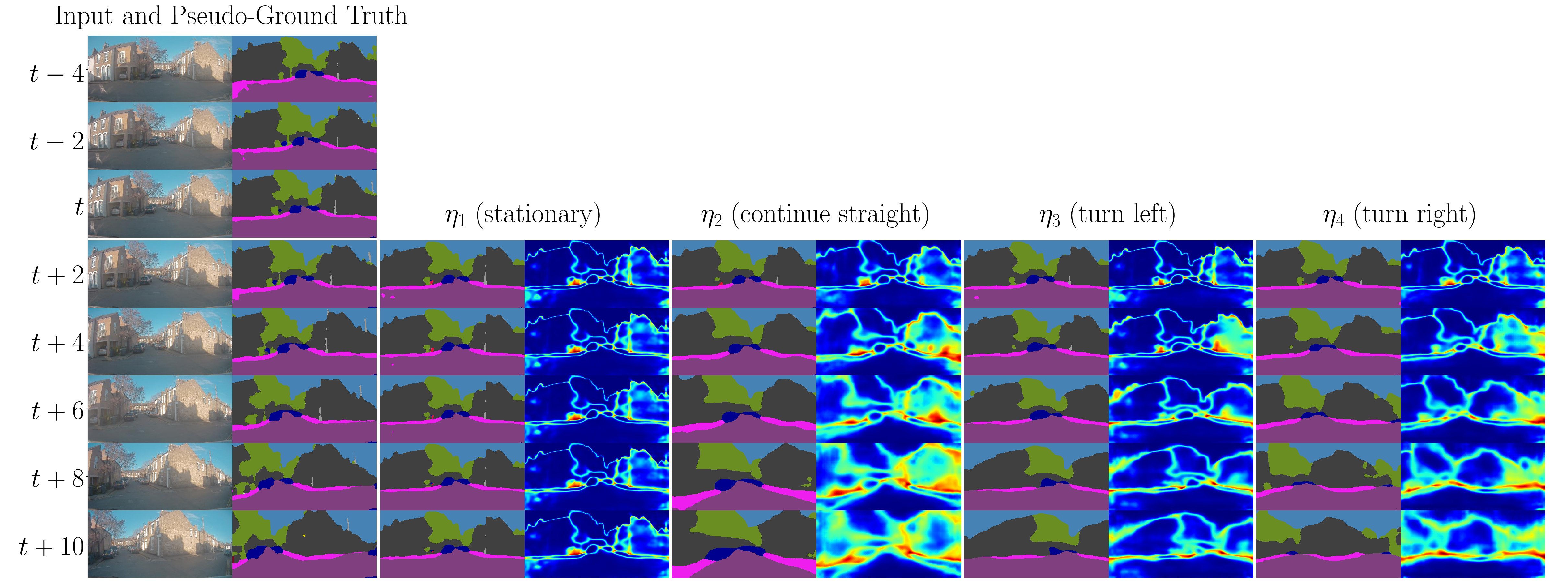}
    \caption{Predicted futures from our model while driving through an urban intersection. From left, we show the actual past and future video sequence and labelled semantic segmentation. Using four different noise vectors, $\eta$, we observe the model imagining different driving manoeuvres at an intersection: being stationary, driving straight, taking a left or a right turn. We show both predicted semantic segmentation and entropy (uncertainty) for each future. This example demonstrates that our model is able to learn a probabilistic embedding, capable of predicting multi-modal and plausible futures.}
    \label{fig:scenarios1}
\end{figure*}

\begin{figure*}[h]
\vspace{-5pt}
    \centering
    \includegraphics[width=\textwidth]{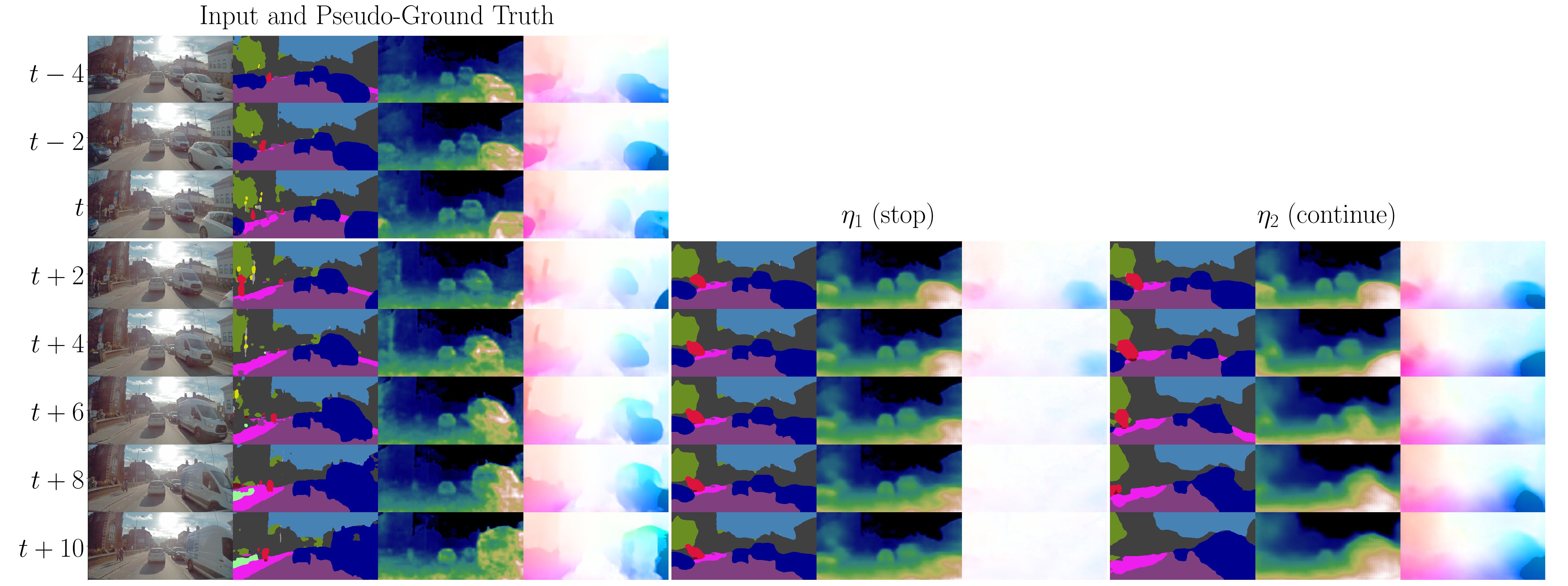}
    \caption{Predicted futures from our model while driving through a busy urban scene. From left, we show actual past and future video sequence and labelled semantic segmentation, depth and optical flow. Using two different noise vectors, $\eta$, we observe the model imagining either stopping in traffic or continuing in motion. This illustrates our model's efficacy at jointly predicting holistic future behaviour of our own vehicle and other dynamic agents in the scene across all modalities.}
    \label{fig:scenarios2}
\vspace{-16pt}
\end{figure*}

Perhaps the most striking result of the model is observing that our model can predict diverse and plausible futures from a single sequence of past frames at $5$Hz, corresponding to one second of past context and two seconds of future prediction.
In \autoref{fig:scenarios1} and \autoref{fig:scenarios2} we show qualitative examples of our video scene understanding future prediction in real-world urban driving scenes.
We sample from the present distribution, $\eta_{t, j} \sim \mathcal{N}(\mu_{t,\text{present}}, \sigma_{t,\text{present}}^2)$, to demonstrate multi-modality.

Further, our framework can automatically infer which scenes are unusual or unexpected and where the model is uncertain of the future, by computing the differential entropy of the present distribution. Simple scenes (e.g. one-way streets) will tend to have a low entropy, corresponding to an almost deterministic future. Any latent code sampled from the present distribution will correspond to the same future. Conversely, complex scenes (e.g. intersections, roundabouts) will be associated with a high-entropy. Different samples from the present distribution will correspond to different futures, effectively modelling the stochasticity of the future.\footnote{In the accompanying \href{https://wayve.ai/blog/predicting-the-future}{blog post}, we illustrate how diverse the predicted future becomes with varying levels of entropy in an intersection scenario and an urban traffic scenario.}

Finally, to allow reproducibility, we evaluate our future prediction framework on Cityscapes \cite{cityscapes16} and report future semantic segmentation performance in \autoref{table:cityscapes}. We compare our predictions, at resolution $256\times512$, to the ground truth segmentation at 5 and 10 frames in the future. Qualitative examples on Cityscapes can be found in \Cref{appendix:cityscapes}.

\begin{table*}[thb!]
\vspace{-16pt}
\centering
\begin{tabular}{ll|cc}
\thickhline
& {\textbf{\color{darkgray}{\shortstack{Temporal Model}}}} & $\text{IoU}_{i=5}$ ($\uparrow$) & $\text{IoU}_{i=10}$  ($\uparrow$) \\
\mediumhline
& Repeat frame & 0.393  & 0.331\\
& Nabavi \etal \cite{nabavi18} & - & 0.274\\
& Chiu \etal \cite{chiu19} & - & 0.408\\
\hline
\multirow{4}{*}{\textbf{\color{darkgray}{\shortstack{Probabilistic \quad}}}} & Res. 3D Conv. \cite{hara17} & 0.445 & 0.399\\
& Conv. GRU \cite{ballas16} & 0.449 & 0.397\\
& Sep. Inception \cite{chen18}& 0.455 & 0.402\\
& \textbf{Ours} & \textbf{0.464} & \textbf{0.416}\\
\thickhline
\end{tabular}
\caption{Future semantic segmentation performance on Cityscapes at $i=5$ and $i=10$ frames in the future (corresponding to respectively 0.29s and 0.59s).}
\label{table:cityscapes}
\vspace{-40pt}
\end{table*}

\subsection{Driving Policy}
We study the influence of the learned temporal representation on driving performance. Our baseline is the control policy learned from a single frame. 

First we compare to this baseline a model that was trained to directly optimise control, without being supervised with future scene prediction. It shows only a slight improvement over the static baseline, hinting that it is difficult to learn an effective temporal representation by only using control error as a learning signal.

\begin{table*}[thb!]
\centering
\begin{tabular}{ll|ccc}
\thickhline
& {\textbf{\color{darkgray}{\shortstack{Temporal Model}}}} & $\mathcal{M}_{\text{perception}} (\uparrow)$ &  Steering ($\downarrow$) & Speed ($\downarrow$) \\
\mediumhline
& Static & - & 0.049 & 0.048\\
& Ours w/o future pred. & - & 0.043 & 0.039\\
\hline
\multirow{4}{*}{\textbf{\color{darkgray}{\shortstack{Deterministic \quad}}}} & Res. 3D Conv. \cite{hara17} & 6.9\% & 0.039 & 0.031\\
& Conv. GRU \cite{ballas16} & 7.4\% & 0.041 & 0.032\\
& Sep. Inception \cite{chen18} & 9.6\% & 0.040 & 0.031\\
& \textbf{Ours} & \textbf{13.6\%} & \textbf{0.036} & \textbf{0.030}\\
\hline
\multirow{4}{*}{\textbf{\color{darkgray}{\shortstack{Probabilistic \quad}}}} & Res. 3D Conv. \cite{hara17} & 8.1\% & 0.040 & 0.028\\
& Conv. GRU \cite{ballas16} & 9.0\% & 0.038 & 0.029\\
& Sep. Inception \cite{chen18} & 13.8\% & 0.036 & 0.029\\
& \textbf{Ours} & \textbf{20.0\%} & \textbf{0.033} & \textbf{0.026}\\
\thickhline
\end{tabular}
\caption{Evaluation of the driving policy. The policy is learned from temporal features explicitly trained to predict the future. We observe a significant performance improvement over non-temporal and non-future-aware baselines.} %We measure mean absolute error of speed and steering outputs, balanced by steering histogram bins and compare these to future prediction perception metrics. 
\label{table:perception-and-control}
\vspace{-16pt}
\end{table*}

All deterministic models trained with the future prediction loss outperform the baseline, and more interestingly the temporal representation's ability to better predict the future (shown by $\mathcal{M}_{\text{perception}}$) directly translate in a control performance gain, with our best deterministic model having, respectively, a 27\% and 38\% improvement over the baseline for steering and speed.

Finally, all probabilistic models perform better than their deterministic counterpart, further demonstrating that modelling the uncertainty of the future produces a more effective spatio-temporal representation. Our probabilistic model achieves the best performance with a 33\% steering and 46\% speed improvement over the baseline.
\section{Conclusions}

% Future work: improve the sharpness of the future predictions using a GAN loss
% Future work: use importance sampling rather than naive sampling within the vae such that we can compute the likelihood of samples
% While the ultimate task we want to master is vehicle control, our results show that having a robust scene representation which contains dynamics is hugely important.
% 

This work is the first to propose a deep learning model capable of probabilistic future prediction of ego-motion, static scene and other dynamic agents.
We observe large performance improvements due to our proposed temporal video encoding architecture and probabilistic modelling of present and future distributions.
% Furthermore, our representation significantly improves the accuracy of a learnt driving policy.
This initial work leaves a lot of future directions to explore: leveraging known priors and structure in the latent representation, conditioning the control policy on future prediction and applying our future prediction architecture to model-based reinforcement learning.

\paragraph{Acknowledgments.} We acknowledge the partial support of Toshiba Europe, grant G100453. We also thank Przemyslaw Mazur, Nikolay Nikolov and Roberto Cipolla for the many insightful research discussions.

%\clearpage
% ---- Bibliography ----
%
% BibTeX users should specify bibliography style 'splncs04'.
% References will then be sorted and formatted in the correct style.
%
\small{
\bibliographystyle{splncs04}
\bibliography{egbib}
}

\clearpage

\appendix

\section{Architecture}
\label{appendix:arch}

In total, our architecture has 30.4M parameters, comprising of modules:
\begin{itemize}
    \item Perception, $E_{\text{perception}}$, 25.3M parameters ;
    \item Dynamics, $\mathcal{Y}$, and present/future distributions, $\mathcal{P}$ and $\mathcal{F}$, 0.8M parameters ;
    \item Future prediction, $\mathcal{G}$, 3.5M parameters ;
    \item Control policy model, $\mathcal{C}$, 0.7M parameters.
\end{itemize}

\subsection{Perception}

\paragraph{Semantics and Geometry.}
Our model is an encoder-decoder model with five encoder blocks and three decoder blocks, followed by an atrous spatial pyramid pooling (ASPP) module \cite{chen17}. The encoders contain 2, 4, 8, 8, 8 layers respectively, downsampling by a factor of two each time with a strided convolution. The decoders contain 3 layers each, upsampling each time by a factor of two with a sub-strided convolution. All layers have residual connections and many are low rank, with varying kernel and dilation sizes. Furthermore, we employ skip connections from the encoder to decoder at each spatial scale.

We pretrain the scene understanding encoder on a number of heterogeneous datasets to predict semantic segmentation and depth: CityScapes \cite{cityscapes16}, Mapillary Vistas \cite{neuhold_mapillary_2017}, ApolloScape \cite{huang2018apolloscape} and Berkeley Deep Drive \cite{yu2018bdd100k}. We collapse the classes to $14$ semantic segmentation classes shared across these datasets and sample each dataset equally during training. We train for 200,000 gradient steps with a batch size of $32$ using SGD with an initial learning rate of $0.1$ with momentum $0.9$. We use cross entropy for segmentation and the scale-invariant loss \cite{li_megadepth_2018} to learn depth with a weight of $1.0$ and $0.1$, respectively.

\paragraph{Motion.}
In addition to this semantics and geometry encoder, we also use a pretrained optical flow network, PWCNet~\cite{sun_pwc-net_2018}. We use the pretrained authors' implementation.

\paragraph{Perception.}
To form our perception encoder we concatenate these two feature representations (from the perception encoder and optical flow net) concatenated together. 
We use the features two layers before the output optical flow regression as the feature representation.
The decoders of these networks are used for generating pseudo-ground truth segmentation and depth labels to train our dynamics and future prediction modules.

\subsection{Training}
Our model was trained on 8 GPUs, each with a batch size of 4, for 200,000 steps using an Adam optimiser with learning rate $3\mathrm{e}{-4}$. The input of our model is a sequence of 15 frames at resolution $224\times480$ and a frame rate of 5Hz ($256\times512$ and 17Hz for Cityscapes). The first 5 frames correspond to the past and present context (1s), and the following 10 frames to the future we want to predict (2s). All layers in the network use batch normalisation and a ReLU activation function. We now describe each module of our architecture in more detail.

\paragraph{Dynamics} four temporal blocks with kernel size $k=(2,3,3)$, stride $s=1$ and output channels $c=[80, 88, 96, 104]$. In between every temporal block, four 2D residual convolutions ($k=(3,3)$, $s=1$) are inserted.

\paragraph{Present and Future Distribution} two downsampling 2D residual convolutions ($k=(3,3)$, $s=2$, $c=[52, 52]$). An average pooling layer flattens the feature spatially, and a final dense layer maps it to a vector of size $2L$ ($L=16$).

\paragraph{Future Prediction} the main structure is a convolutional GRU ($k=(3,3)$, $s=1$). Each convolutional GRU is followed by three 2D residual convolutions ($k=(3,3)$, $s=1$). This structure is stacked five times.
The decoders: two upsampling convolutions ($k=(3,3)$, $s=1$, $c=32$), a convolution ($k=(3,3)$, $s=1$, $c=16$), and finally a convolution without activation followed by a bilinear interpolation to the original resolution $224\times480$.

\paragraph{Control} two downsampling convolutions ($k=(3,3)$, $s=2$, $c=[64, 32]$), followed by dense layers ($c=[1024, 512, 256, 128, 64, 32, 16, 4]$).

\subsection{Temporal Block}
\label{appendix:temporal-block}
We ablate the architecture of our proposed Temporal Block module on Cityscapes by evaluating performance of future semantic segmentation prediction, at resolution $256 \times 512$ and for future frames 5 and 10. Let $k_t$ denote the temporal kernel size and $k_s$ the spatial kernel size of the 3D convolutions. We compare the following modules:

\begin{enumerate}[label=(\roman*)]
    \item ($k_t, k_s, k_s$) and ($1, k_s, k_s$) convolutions. No global context.
    \item ($k_t, k_s, k_s$), ($k_t, 1, k_s$) and ($1, k_s, k_s$) convolutions. No global context.
    \item ($k_t, k_s, k_s$), ($k_t, k_s, 1$) and ($1, k_s, k_s$) convolutions. No global context.
    \item ($k_t, k_s, k_s$), ($k_t, 1, k_s$), ($k_t, k_s, 1$) and ($1, k_s, k_s$) convolutions. No global context.
    \item ($k_t, k_s, k_s$), ($k_t, 1, k_s$), ($k_t, k_s, 1$) and ($1, k_s, k_s$) convolutions. With global context (\ie our proposed Temporal Block).

\end{enumerate}

\begin{table*}[thb!]
\centering
\begin{tabular}{ll|cc}
\thickhline
& {\textbf{\color{darkgray}{\shortstack{Temporal Model}}}} & $\text{IoU}_{i=5}$ ($\uparrow$) & $\text{IoU}_{i=10}$  ($\uparrow$) \\
\mediumhline
& Repeat frame & 0.393  & 0.331\\
\hline
\multirow{5}{*}{\textbf{\color{darkgray}{\shortstack{Probabilistic \quad}}}} & (i) ($k_t, k_s, k_s$) & 0.454 & 0.411\\
& (ii) ($k_t, k_s, k_s$), ($k_t, 1, k_s$) & 0.461  & 0.411\\
& (iii) ($k_t, k_s, k_s$), ($k_t, k_s, 1$)& 0.449 & 0.413\\  
& (iv) ($k_t, k_s, k_s$), ($k_t, 1, k_s$), ($k_t, k_s, 1$) ~& 0.453& 0.413\\
& (v) \textbf{Temporal Block (Ours)} & \textbf{0.464} & \textbf{0.416}\\
\thickhline
\end{tabular}
\caption{Ablation study of the Temporal Block on Cityscapes, evaluated on future semantic segmentation performance at $i=5$ and $i=10$ frames in the future. Our proposed Temporal Block module outperforms all the other variants.}
\vspace{-8pt}
\end{table*}

\clearpage
\section{Nomenclature}
\label{appendix:nomenclature}

We detail the symbols used to describe our model in this paper.

% So that table is at the top of the page.
\makeatletter
\setlength{\@fptop}{0pt}
\makeatother

\begin{table}[!htb]
\centering
\resizebox{\columnwidth}{!}{
\begin{tabular}{l|l}
    \multicolumn{2}{l}{\textbf{Networks}}\\
    \hline
    Perception encoder &  $E_{\text{perception}}$ \\
    Temporal Block & $\mathcal{T}$ \\
    Dynamics module & $\mathcal{Y}$ \\
    Present network & $\mathcal{P}$ \\
    Future network & $\mathcal{F}$ \\
    Future prediction module & $\mathcal{G}$ \\
    Future decoders & $\mathcal{D}_s, \mathcal{D}_d, \mathcal{D}_f$ \\
    Control module & $\mathcal{C}$ \\
    \hline
    
    \multicolumn{2}{l}{\textbf{Tensors}}\\
    \hline
    Temporal context & $T$ \\
    Future prediction horizon & $N_f$ \\
    Future control horizon & $N_c$ \\
    Input image & $i_t$ \\
    Perception features & $x_t = E_{\text{perception}}(i_t)$ \\
    Dynamics features & $z_t = \mathcal{Y}(x_{t-T+1}:x_t)$ \\
    Present distribution & $\mu_{t,\text{present}}, \sigma_{t,\text{present}} = \mathcal{P}(z_t)$ \\
    Future distribution & $\mu_{t,\text{future}}, \sigma_{t,\text{future}} = \mathcal{F}(z_t)$ \\
    Noise vector (train) & $\eta_t \sim \mathcal{N}(\mu_{t,\text{future}}, \sigma_{t,\text{future}}^2)$ \\
    Noise vector (test) & $\eta_t \sim \mathcal{N}(\mu_{t,\text{present}}, \sigma_{t,\text{present}}^2)$ \\
    Future prediction inputs & $u_t^{t+i} = \eta_t$\\
    Future prediction initial hidden state & $g_t^{t} = z_t$\\
    Future prediction output features & $g_t^{t+i} = \mathcal{G}(u_t^{t+i}, g_t^{t+i-1})$\\
    $\begin{aligned} \text{Future perception outputs}  \\ \end{aligned}$ & $\begin{aligned}\hat{o}_t^{t+i} &= \{\hat{s}_t^{t+i}, \hat{d}_t^{t+i}, \hat{f}_t^{t+i} \}  \\
    & = \{\mathcal{D}_s(g_t^{t+i}), \mathcal{D}_d(g_t^{t+i}), \mathcal{D}_f(g_t^{t+i}) \} \end{aligned}$\\
    $\begin{aligned} \text{Control outputs}  \\ \end{aligned}$ & $\begin{aligned}\hat{c}_t  &= \{\hat{v}_t, \hat{\dot{v}}_t, \hat{\theta}_t, \hat{\dot{\theta}}_t\}  \\
    & = \mathcal{C}(z_t) \end{aligned}$\\
\end{tabular}}
\end{table}

\clearpage
\section{Cityscapes Qualitative Examples}
\label{appendix:cityscapes}
\captionsetup[subfigure]{justification=justified,singlelinecheck=false}
\begin{figure}[h]
    \vspace{-10pt}
    \centering
    \begin{subfigure}[b]{\textwidth}
        \centering
        \includegraphics[width=0.7\linewidth]{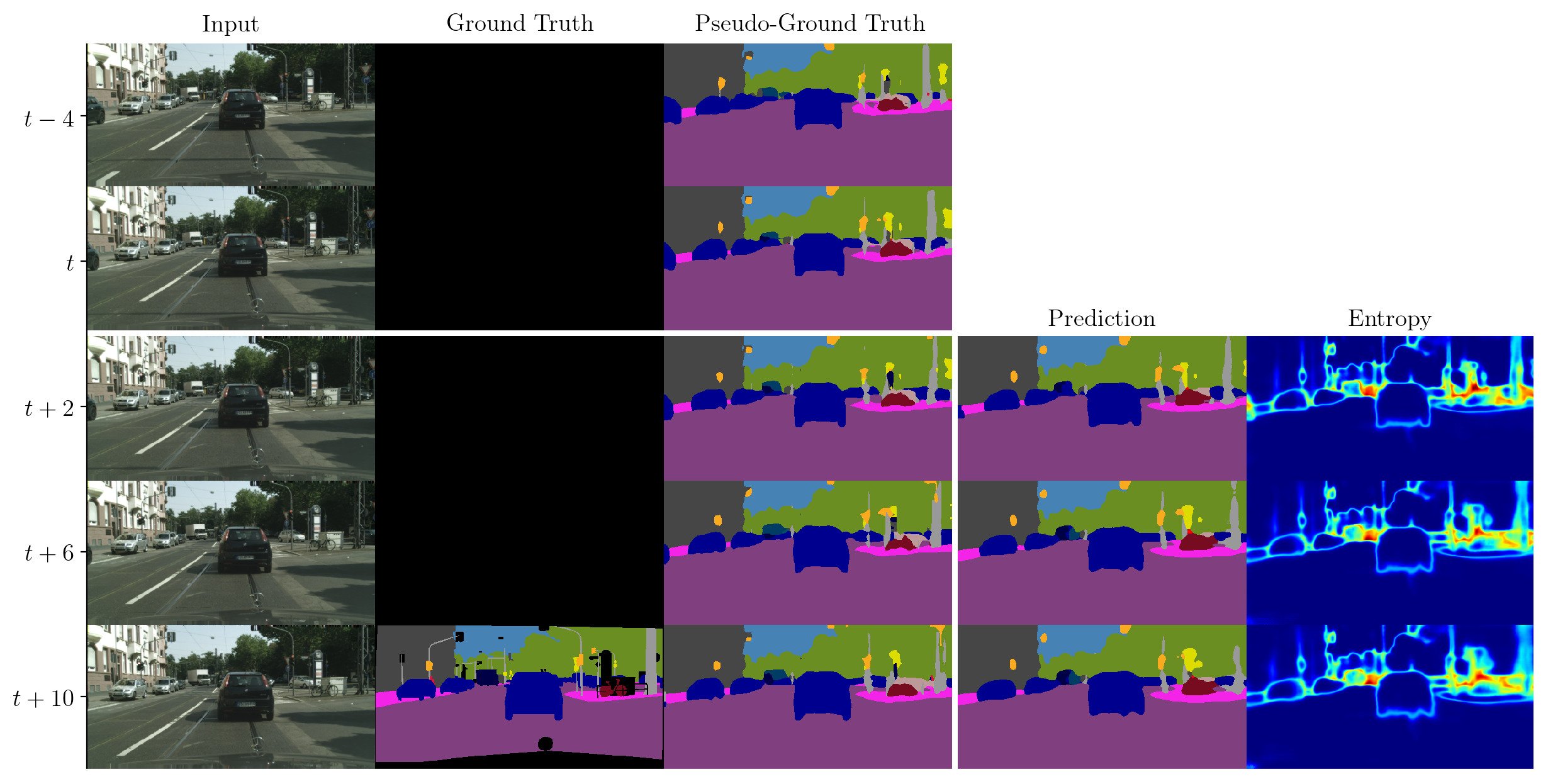}
        \caption{Our model can correctly predict future segmentation of small classes such as poles or traffic lights.}
        \label{fig:partial-occlusion}
    \end{subfigure}
    \vskip\baselineskip
    \begin{subfigure}[b]{\textwidth}
        \centering
        \includegraphics[width=0.7\linewidth]{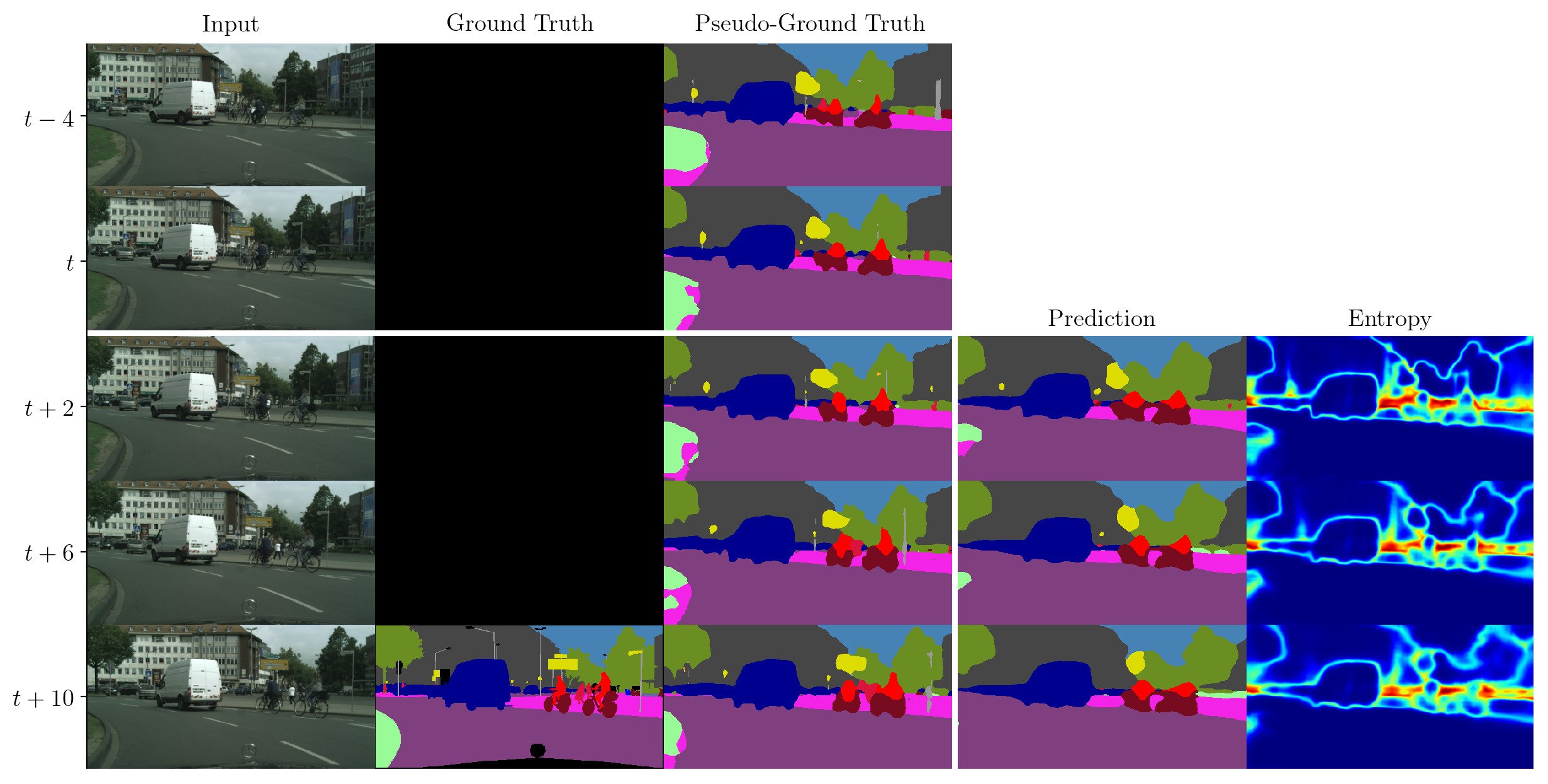}
        \caption{Dynamic agents, \ie cars and cyclists, are also accurately predicted.}
        \label{fig:continuous-tracking}
    \end{subfigure}
    \vskip\baselineskip
    \begin{subfigure}[b]{\textwidth}
        \centering
        \includegraphics[width=0.7\linewidth]{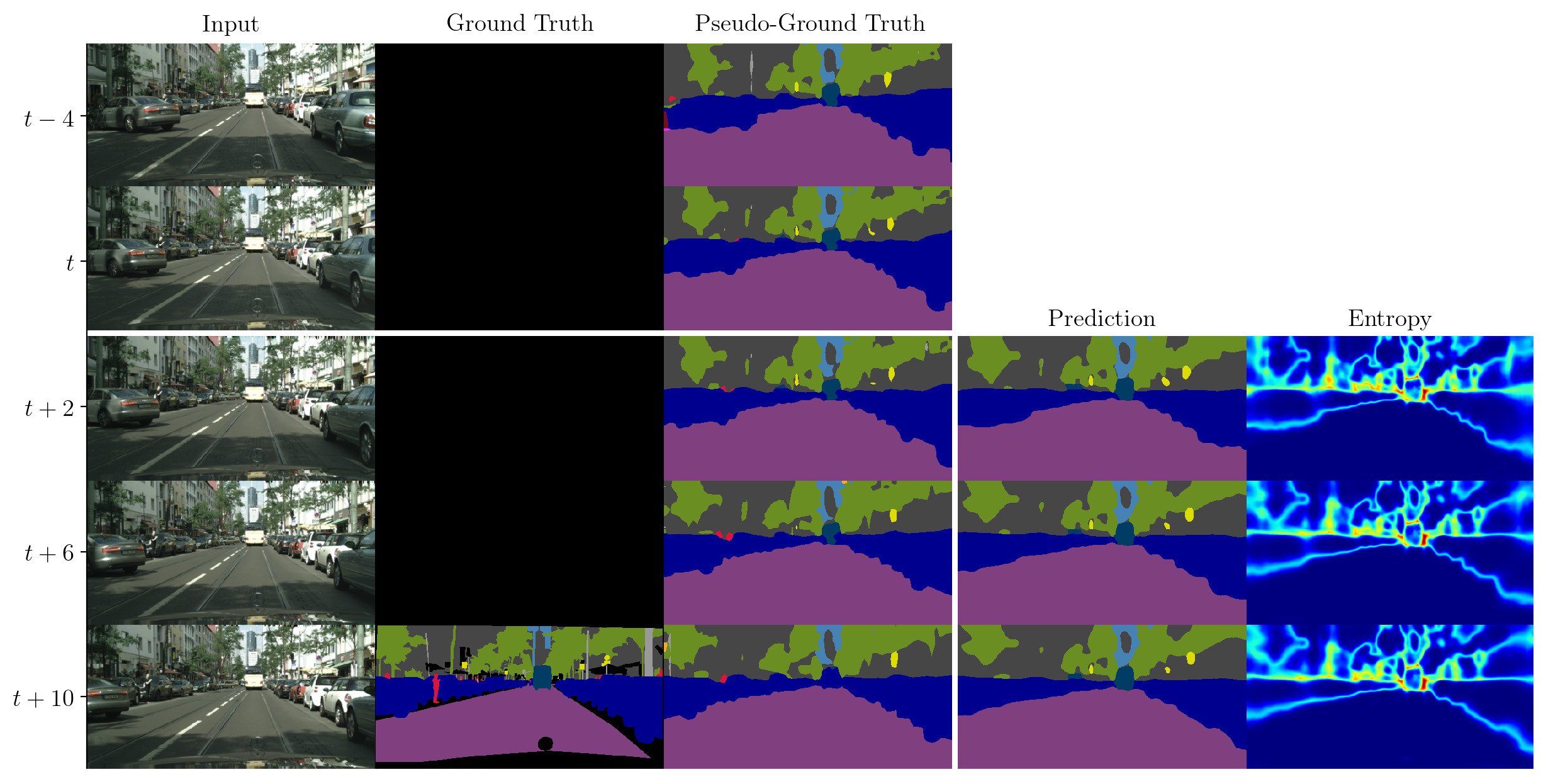}
        \caption{In this example, the bus is correctly segmented, without any class bleeding contrary to the pseudo-ground truth segmentation, showing that our model can reason in a holistic way.}
        \label{fig:total-occlusion}
    \end{subfigure}
    \caption{Future prediction on the CityScapes dataset, for 10 frames in the future at 17Hz and $256\times 512$ resolution.}
    \vspace{-110pt}
\end{figure}

\end{document}